\documentclass[conference]{IEEEtran}
\IEEEoverridecommandlockouts
\usepackage{cite}
\usepackage{amsmath,amssymb,amsfonts}
\usepackage{algorithmic}
\usepackage{graphicx}
\usepackage{textcomp}
\usepackage{xcolor}
\usepackage[export]{adjustbox}
\usepackage[bookmarks=false]{hyperref}
\usepackage{footnote}
\usepackage{caption}
\usepackage{subcaption}

\usepackage[utf8]{inputenc}
\usepackage[english]{babel}

\usepackage{siunitx}
\ExplSyntaxOn
  \cs_new_eq:NN \calc \fp_eval:n
\ExplSyntaxOff
\newcommand*{\formatNumber}[2][]{\num[%
  round-mode=places,% Round output to specified number of places
  round-precision=4,% Round-precision is 4
  output-decimal-marker={.},% Use . as decimal marker
  #1% Other options
  ]{\calc{#2}}}
\def\BibTeX{{\rm B\kern-.05em{\sc i\kern-.025em b}\kern-.08em
    T\kern-.1667em\lower.7ex\hbox{E}\kern-.125emX}}

\begin{document}

\title{Illumination-Based Data Augmentation for Robust Background Subtraction\\
}

\author{\IEEEauthorblockN{Dimitrios Sakkos\IEEEauthorrefmark{1} \thanks{\IEEEauthorrefmark{1} email: dksakkos@gmail.com},
Hubert P. H. Shum\IEEEauthorrefmark{2} \thanks{\IEEEauthorrefmark{2} email: hubert.shum@northumbria.ac.uk} and
Edmond S. L. Ho\IEEEauthorrefmark{3} \thanks{\IEEEauthorrefmark{3} email: e.ho@northumbria.ac.uk, \textit{corresponding author}}}
\IEEEauthorblockA{\textit{Department of Computer and Information Sciences} \\
\textit{Northumbria University}\\
Newcastle upon Tyne, United Kingdom \\
}
}

\IEEEoverridecommandlockouts

\maketitle

\IEEEpubidadjcol

\begin{abstract}
A core challenge in background subtraction (BGS) is handling videos with sudden illumination changes in consecutive frames. In this paper, we tackle the problem from a data point-of-view using data augmentation. 
%propose a method that addresses the limitation of traditional data augmentation approaches: 
Our method performs data augmentation that not only creates endless data on the fly, but also features semantic transformations of illumination which enhance the generalisation of the model. It successfully simulates flashes and shadows by applying the Euclidean distance transform over a binary mask generated randomly. 
Such data allows us to effectively train an illumination-invariant deep learning model for BGS.
%We further propose a post-processing method that removes noise from the output binary map of segmentation, resulting in a cleaner, more accurate segmentation map.
%that can generalise to multiple scenes of different conditions. 
%We show that is is possible to train deep learning models even with very limited training samples. 
Experimental results demonstrate the contribution of the synthetics in the ability of the models to perform BGS even when significant illumination changes take place. The source code of the project is made publicly available at \url{https://github.com/dksakkos/illumination_augmentation}.
\end{abstract}


\section{Introduction}
Background subtraction (BGS) has been an active research area in the past decades. The main task is to differentiate the foreground (i.e. moving objects) from the background (i.e. the static parts of a given scene) \cite{Barnich2011}. A large number of real-world applications, such as person re-identification \cite{ReID_BGS}, object tracking \cite{target_tracking_BGS}, gesture recognition \cite{gesture_recognition_BGS}, vehicle tracking \cite{vehicle_tracking_1_BGS},  video recognition \cite{li_ming_2019}, action recognition \cite{Ho:2015:MSR:2821592.2821617,Ho2016}, crowd analysis \cite{crowd_analysis_BGS} and even use cases of the medical domain \cite{sensitivity_enhancement_BGS,Mccay:EMBC}, depend on accurate and robust background subtraction as a first step of their pipelines. 

Sudden illumination changes signify a particularly difficult challenge, since they cannot be captured by a background model. Such changes in lighting conditions can be caused either by weather conditions or electric lights and result in color changes of a significant amount of pixels.  Due to the difference of visual appearance in consecutive frames, BGS becomes inaccurate. The timing of these changes could be short, such as switching a light on/off, or a piece of cloud blocking the sun, making it tough for the system to adjust to the new condition in a timely manner. 
%In addition, a scene and the objects that appear in it will drastically transform during the night. It is necessary for an algorithm to be able to adjust in this kind of conditions.

State-of-the-art deep learning algorithms allow adapting to sudden illumination changes if a huge amount of training data is provided. However, obtaining labelled data is very costly and there is only limited datasets available in the community. As a solution, data augmentation methods are proposed to perform image-based operations on the data, such as mirroring or cropping, to synthesize a larger dataset. However, simple image tricks cannot effectively generate images with realistic illumination changes. Another solution is adding a small amount of noise to create a new, synthetic image that is similar to the original in context but different in color distribution. %A major advantage is that each synthetic image will be unique, due to the added noise being random. However, the downside is that the added noise does not have any semantic meaning. Therefore, although the synthetic images do increase the generalisation of the model simply by obscuring pixels of the original image, they do not offer any additional knowledge of different circumstances of the same scene.
However, since the added noise does not have any semantic meaning, the synthetic images only slightly increase the generalisation power of the model, as they do not offer any additional knowledge of different lighting conditions of the same scene.

%To overcome this challenge, we propose a novel image augmentation method that generates synthetic images by altering the illumination of the input in both local areas and the global image. Such agumented data allows us to provide extra semantic information to the BGS model in terms of illumination, which leads to better generalisation in scenes depicting light-based effects such as halos and shadows. We show that this approach yields significantly better results compared to traditional augmentation techniques, particularly in data sets that feature changes of brightness. We further improve the system with a post-processing method that takes advantage of the temporal information of the input video. In particular, we exploit the limited movement of the foreground objects between adjacent frames, in order to filter noise caused by false positives in some parts of the image.

To overcome this challenge, we propose a new data augmentation technique by synthesising the light-based effects of different degrees of brightness. Such effects include shadows and halos of different size, placed in random locations of the input image. In addition, global illumination changes are also included, in order to increase the generalisation abilities of the model to scenes filmed at various times of the day and night. Such augmented data allows us to provide extra semantic information to the BGS model in terms of illumination for better generalisation performance. The results show that the proposed technique is superior to regular augmentation methods and can significantly boost the segmentation results even in scenes that feature illumination conditions unseen to the model. Our experiments indicate that the proposed method improves the BGS results in our quantitative and qualitative evaluations on the benchmark dataset SABS \cite{sabs}.

The main contributions of this work can be summarized as follows:
\begin{itemize}
    \item A novel synthetic image generation method for robust background subtraction under challenging illumination conditions.
    \item An illumination-invariant deep neural network for background subtraction.
\end{itemize}

This  paper  is  organised  as  follows. A literature review on the task of background subtraction is given in Section \ref{sec:related_work}. Section \ref{sec:methodology} outlines the proposed method for synthetic generation covering local, global and combined changes. In Section \ref{sec:experimental_settings} we introduce the dataset and we describe the training settings of our models. Section  \ref{sec:results} follows with the presentation of our results and discussion. Finally, Section \ref{sec:conclusion} provides the conclusion and future work.

\section{Related Work \label{sec:related_work}}
In this section, we will first review the related research in background subtraction using traditional approaches such as Gaussian Mixture Models and Principal Component Analysis. Then, we discuss on recent deep-learning based research. 

Performing background subtraction on video with illumination change has been explored in the literature. In particular, Siva et al. \cite{siva_shafiee_li_wong_2015} demonstrated that the pixel intensity values affected by sudden local illumination change can be modelled by combining a GMM with a conditional probabilistic function based on an extension of Zivkovic et al. \cite{Zivkovic2006}. Akilan et al. \cite{akilan_wu_yang_2018} proposed a feature fusing approach to fuse the color distortion, color similarity, and illumination measures to improve the performance of the GMM-based model. Vosters et al. \cite{vosters} presented a statistical illumination model based on a PCA-based approach \cite{Oliver}. Such an approach is similar to Pillet et al. \cite{pillet} in which a spatial likelihood model for modelling the relationship of neighboring pixels is proposed. When new frames are analysed, the model is updated. While the aforementioned approaches tried to improve the performance of background subtraction on videos with illumination change, the GMM-based approaches fail when there is a significant illumination change in consecutive frames \cite{BGS_review_2014}. PCA-based approaches are more robust in handling illumination changes in general. However, the lack of semantic knowledge in the scene limits the performance of PCA-based approaches.

In the past few years, the area of computer vision has grown rapidly thanks to deep learning. Deep neural networks are now the best performing models in background subtraction considering accuracy and robustness. This task's primary objective is to perform binary pixel-wise foreground/background classification in a given image or video. Clearly, since pixel-wise precision is the target, attention to detail is required. Similar to our proposed methods, there is existing work taking advantage of spatio-temporal information to improve the performance. A 3D convolution-based approached is proposed by Sakkos et al. \cite{Sakkos2018} to exploit the relationship between a block of 10-frame for background subtraction tasks. In \cite{berjon_2018}, the background model of the Kernel Density Estimation-based system is updated using information from previous frames. Group property information is exploited in both spatial and temporal domains in the sparse signal recovery based approached proposed by Liu et al \cite{LiuBG}. A recent work \cite{B-SSSR} further demonstrated incorporating spatio-temporal constraints to improve \cite{rpca1} results in better performance. A successful deep-learning system requires a huge amount of training data, and it is costly to obtain labelled data with significant illumination variation. In this paper, we focus on a data augmentation approach to create synthetic data for training an illumination-invariant BGS network.

\section{Methodology}
\label{sec:methodology}
In this section, we explain how we synthesise images of different illumination with both local and global changes, and then combine them as a unified augmentation method that covers all scenarios simultaneously. 
%In the following subsections, we explain in detail the creation of the synthetic images for all cases. In addition, we describe the post-processing method for the output refinement.

\subsection{Local changes}
To synthesize local changes of illumination, we generate the synthetic images by locally altering the illumination of the input image, therefore creating either a "lamp-post" light source or a shadow effect. First, we randomly select a pixel of the image that serves as the centre of the illumination circle to be drawn: $p=I(w,h), w \in W, h \in H, I=W\times H$, where $W$, $H$ the width and height of the input image $I$ respectively.
Once the coordinates of the centre pixel are determined, we randomly select the diameter $d$ of the illumination circle. Since we want our model to be robust to both small and large shadows and flashes of light, we choose the diameter to be between one fifth and half of the smallest dimension of the input image: $d=k \times \min (W, H), k \in ({1 \over 5}, {1 \over 2}).$

Since modifying all pixels within the circle uniformly generate unrealistic results, we proposed a more sophisticated approach to model the effect of the light. First, we calculate the binary mask $M_1$ of the pixels to be altered using the following formula:
\begin{equation}
    M_1(x,y)=1 \Leftrightarrow (x-w)^2+(y-h)^2\leq d^2
    \label{eq:M1}
\end{equation}
This means that the pixels of our mask have the value of 1 if they reside within the drawn circle and zero everywhere else.
We then use the Euclidean Distance Transform (EDT) to model the light attenuation. Given a binary mask $B$, EDT is defined as:
\begin{equation}
    EDT_x(B)=min_b(||x-b||_{L_2}),  \quad \forall b \in B,
    \label{eq:edt}
\end{equation}
where $L_2$ is the Euclidean norm.
Now, we can calculate the mask for local changes $M_2$ by applying the EDT on $M_1$: 
\begin{equation}
    M_2=EDT(M_1)
    \label{eq:M2}
\end{equation}

\begin{figure*}
     \centering
    \begin{subfigure}[t]{0.23\textwidth}
         \centering
         \includegraphics[width=1\textwidth]{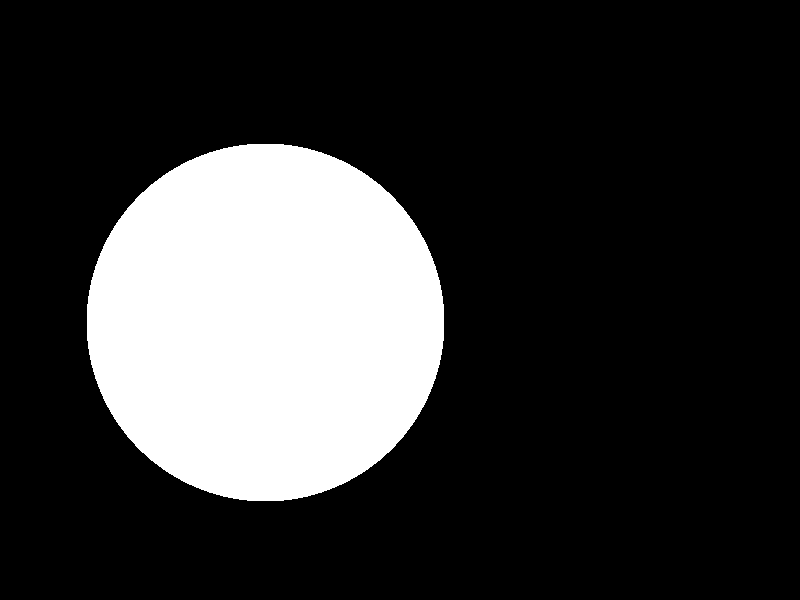}
         \caption{The mask $M_1$}
         \label{fig:M1}
     \end{subfigure} 
     \hfill
     \begin{subfigure}[t]{0.23\textwidth}
         \centering
         \includegraphics[width=1\textwidth]{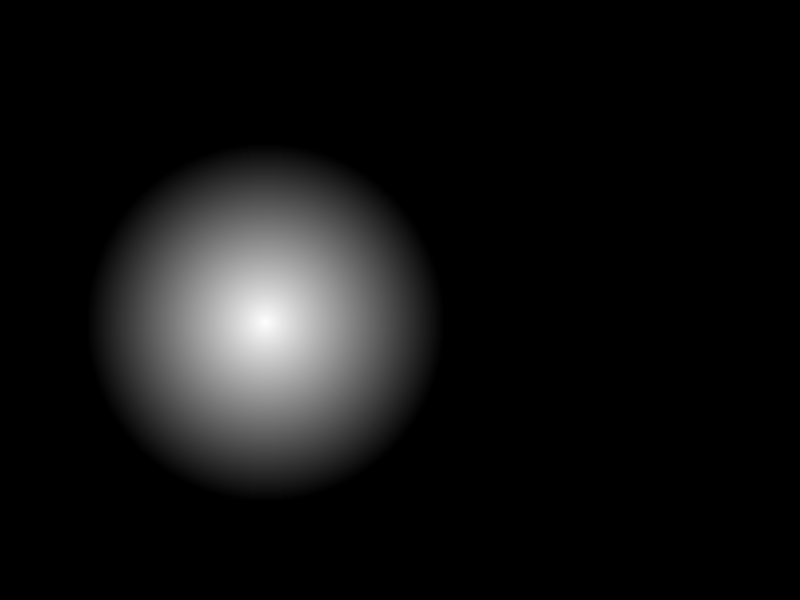}
         \caption{The mask $M_2$}
         \label{fig:M2}
     \end{subfigure}
     \hfill
     \begin{subfigure}[t]{0.23\textwidth}
         \centering
         \includegraphics[width=1\textwidth]{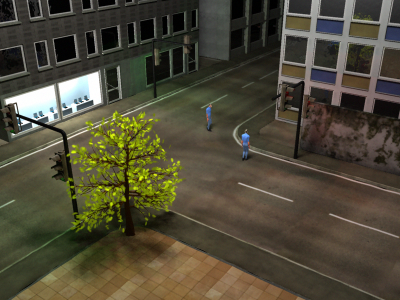}
         \caption{Original image}
         \label{fig:Original}
     \end{subfigure}
     \hfill
     \begin{subfigure}[t]{0.23\textwidth}
         \centering
         \includegraphics[width=1\textwidth]{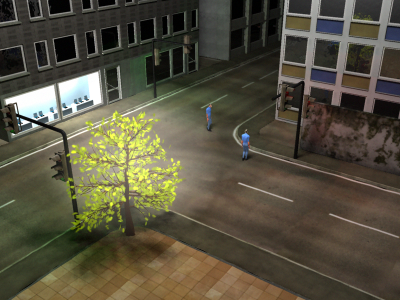}
         \caption{After effect}
         \label{fig:Mres}
     \end{subfigure}
      \caption{The application of the mask for local changes. Subfigure (a): the initial binary mask $M_1$ is created by a circle of diameter $d=179$ and centre coordinates $(322,265)$. Subfigure (b): The mask $M_2$ after the application of the Euclidean distance transform on $M_1$. Subfigures (c) and (d) depict the original image and the lamp-post light source effect after the application of the mask $M_2$ on the input image respectively.}
    \label{fig:local_masks}
\end{figure*}

Once the new mask has been created, we proceed to alter the pixels of the original image that lie within the circle. The new synthetic image $I_s$ is calculated as:
\begin{equation}
    I_s=I\pm(M_2 \times z), \quad z \in [120,160], 
    \label{eq:local}
\end{equation}
where $I$ the original image,
$M_2$ the mask calculated with the distance transform,
$z$ a random integer, and
$\pm$ is either pixel-wise addition or subtraction, chosen with probability $p=0.5$. When the addition operation is chosen, a lamp-post effect will be created in a random part of the image. Conversely, the subtraction operation creates shadows. The application of the aforementioned local masks are depicted in Figure \ref{fig:local_masks}. It can be seen that the final light source effect looks realistic.

\subsection{Global changes}
In some cases, global illumination changes can occur. For example, a lightning during a storm may instantly increase the brightness, and once the rain is over the global illumination will change again. In order to model such illumination changes, we need to alter the pixels of the whole image, rather than a small patch.

We synthesize global illumination changes as:
\begin{equation}
    I_s=I \pm z, \quad z \in [40,80],
    \label{eq:global}
\end{equation}
where $I$, $z$ and $\pm$ are as previously defined. In this case the illumination noise $z$ needs to be slightly diminished, since the whole image is affected.

\subsection{Combined changes}
To capture both local and global illumination changes in the scene, we combine equation \ref{eq:local} and equation \ref{eq:global} into the following:
\begin{eqnarray}
    I_s = z_1\pm(I\pm (M_2 \times z_2), z_1 \in [40, 80], z_2 \in [120,160] 
    \label{eq:unified}
\end{eqnarray}

\begin{figure*}
     \centering
     \begin{subfigure}[t]{0.23\textwidth}
         \centering
         \includegraphics[width=1\textwidth]{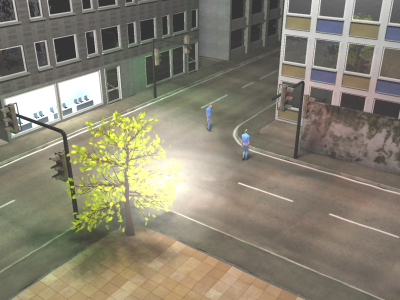}
         \caption{$G_b L_b$}
         \label{fig:bright_bright}
     \end{subfigure} %
     \hfill
     \begin{subfigure}[t]{0.23\textwidth}
         \centering
         \includegraphics[width=1\textwidth]{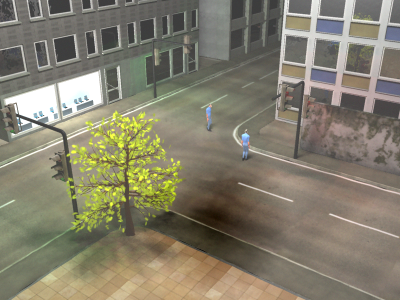}
         \caption{$G_b L_d$}
         \label{fig:bright_dark}
     \end{subfigure} %
     \hfill
     \begin{subfigure}[t]{0.23\textwidth}
         \centering
         \includegraphics[width=1\textwidth]{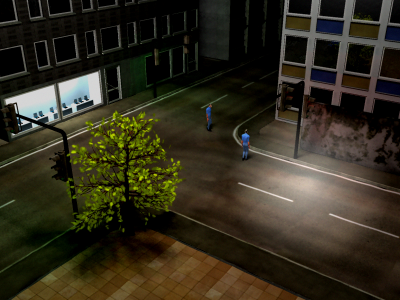}
         \caption{$G_d L_b$}
         \label{fig:dark_bright}
     \end{subfigure} %
     \hfill
     \begin{subfigure}[t]{0.23\textwidth}
         \centering
         \includegraphics[width=1\textwidth]{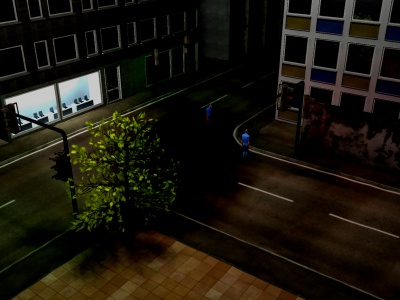}
         \caption{$G_d L_d$}
         \label{fig:dark_dark}
     \end{subfigure}
        \caption{Combination of global and local illumination changes. The subfigures (a) and (b) depict a combination of a brightening global filter with a bright and dark local filter respectively. On the other hand, subfigures (c) and (d) implement the darkening global filter.}
        \label{fig:global_local}
\end{figure*}

Sample images synthesised from our system can be found in figure \ref{fig:global_local}. Since both the positioning and the intensity of the masks is random, this method can effectively cover all kinds of illumination changes. Additionally, hundreds of different synthetic images can be generated from a single frame. Therefore, given a small video, we can generate enough unique synthetic images to train a very deep network.

\subsection{Illumination-invariant Deep Networks}
We utilise the synthetic images to train multiple deep learning networks for BGS and evaluate their performances. Due to the use of images with synthetic illumination changes, these networks become invariant to lighting conditions.

\begin{figure*}
     \begin{subfigure}[b]{0.58\textwidth}
         \centering
         \includegraphics[width=\textwidth]{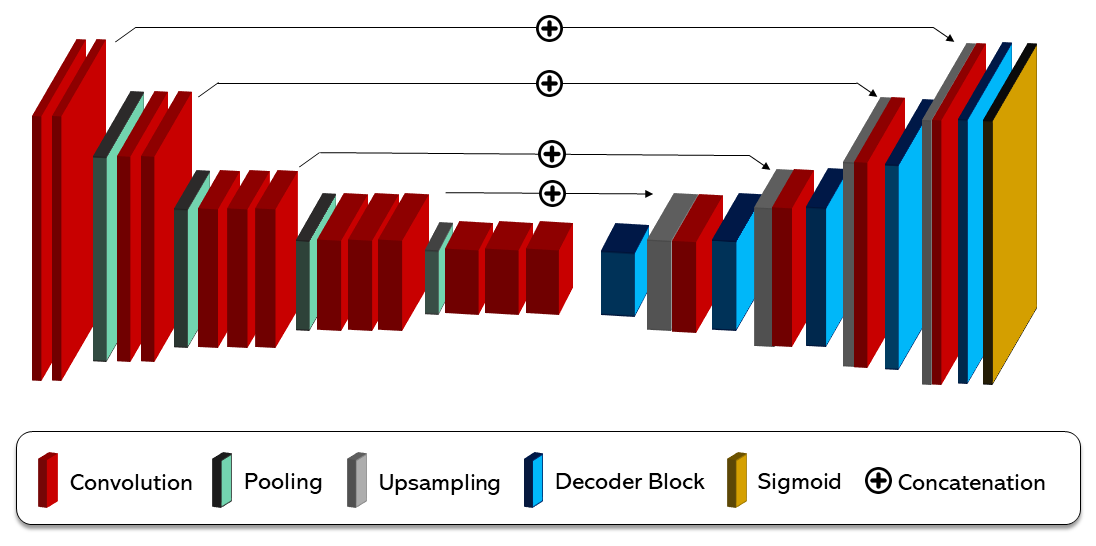}
        \caption{Overview}
     \end{subfigure}
     \hfill
     \begin{subfigure}[b]{0.38\textwidth}
         \centering
         \includegraphics[width=\textwidth]{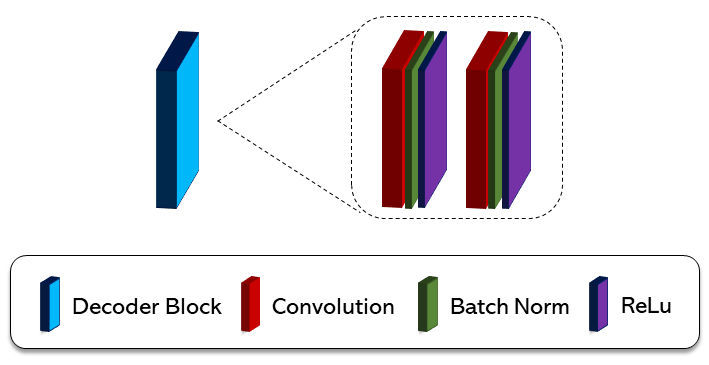}
     \caption{Decoder block}
    \label{fig:decoder_block}
     \end{subfigure}
        \caption{The CNN that was used for the experiments. The encoder is initialised from VGG16 \cite{VGG16} and is keep fixed during training.}
        \label{fig:net_arch}
\end{figure*}

To ensure the fairness of our experiments, all models used in this study have the same architecture. We follow the paradigm of previous background subtraction approaches \cite{FgSegNet, zeng_zhu_2018_} and use a Unet architecture, which comprises an encoder and a decoder. We employ transfer learning and use the VGG16 model \cite{VGG16} as the encoder. Therefore, the weights of the encoder are initialised from those of VGG16, which has been pre-trained on Imagenet. VGG16 encompasses 13 convolutional layers, 5 pooling layers and 3 fully connected layers. Following the work of Long et al. \cite{fcn}, we remove the fully connected layers and make the network fully convolutional. Because of the pooling layers, the output of the encoder is 5 times smaller than the input. We use the decoder to recover the information that is lost from the downsampling operation via the use of upsampling blocks. Each block consists of a 2x2 bilinear interpolation operation which upsamples the feature maps, followed by two 3x3 convolutional layers with batch normalisation applied in-between (Figure \ref{fig:decoder_block}). To maximise the information recovered by the encoder, we add skip connections that connect the encoder to the decoder. In addition, the ReLu non-linearity is applied after each convolutional layer. Finally, once the spatial size has been restored, we add a final 3x3 convolutional layer, followed by a sigmoid layer to convert the output of the model to a foreground probability map. The architecture of the network is illustrated in Figure \ref{fig:net_arch}.

\section{Experiment settings \label{sec:experimental_settings}}

\subsection{Dataset}
In order to highlight the robustness of our augmentation process, we select the Stuttgart Artificial Background Subtraction dataset (SABS) \cite{sabs}. 

\begin{figure}
     \begin{subfigure}[b]{0.15\textwidth}
         \centering
         \includegraphics[width=\textwidth]{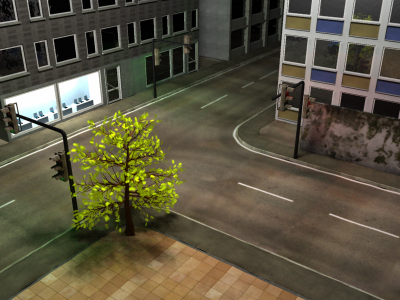}
     %    \caption{$G_b L_b$}
         \label{fig:darkening_start}
     \end{subfigure}
     \hfill
     \begin{subfigure}[b]{0.15\textwidth}
         \centering
         \includegraphics[width=\textwidth]{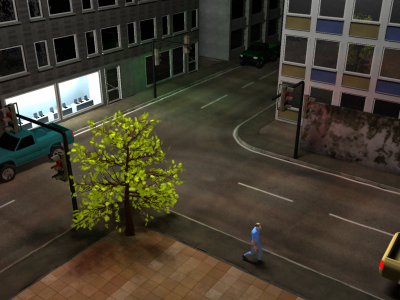}
    %     \caption{$G_b L_d$}
         \label{fig:darkening_med}
     \end{subfigure}
    \hfill
     \begin{subfigure}[b]{0.15\textwidth}
         \centering
         \includegraphics[width=\textwidth]{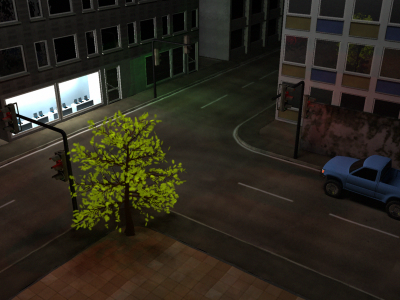}
    %     \caption{$G_d L_b$}
         \label{fig:darkening_end}
     \end{subfigure}
    \\[2ex]
     \begin{subfigure}[b]{0.15\textwidth}
         \centering
         \includegraphics[width=\textwidth]{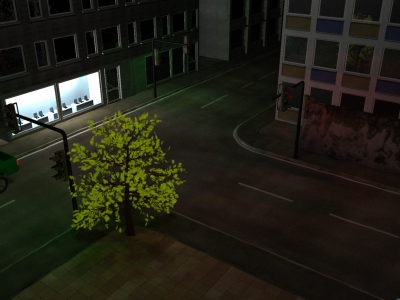}
   %      \caption{$G_b L_b$}
         \label{fig:lightswitch_start}
     \end{subfigure}
     \hfill
     \begin{subfigure}[b]{0.15\textwidth}
         \centering
         \includegraphics[width=\textwidth]{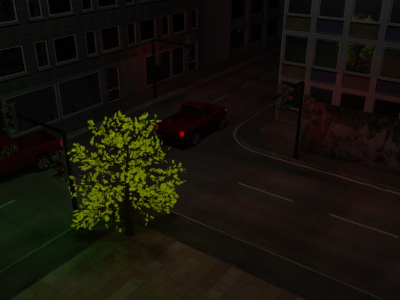}
   %      \caption{$G_b L_d$}
         \label{fig:lightswitch_med}
     \end{subfigure}
    \hfill
     \begin{subfigure}[b]{0.15\textwidth}
         \centering
         \includegraphics[width=\textwidth]{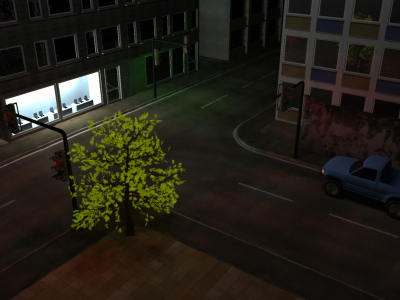}
     %    \caption{$G_d L_b$}
         \label{fig:lightswitch_end}
     \end{subfigure}
        \caption{The SABS dataset that was used for evaluating the models. The first row depicts the training sequence \textit{Darkening}, while the second row shows the testing video \textit{LightSwitch}. The columns show frames from the start, middle and ending parts of the video. Note that in the middle of the \textit{LightSwitch} sequence the store light switches off, causing major changes to the background.}
        \label{fig:SABS_overview}
\end{figure}

The SABS dataset \cite{sabs} contains 9 synthetic video sequences. The main challenge of the dataset stems from the sudden change of illumination over time. Although the foreground movements are the same in some sequences, the illumination is changing over time. In addition, different videos have very different lighting conditions, such as day-time and night scenes. In our experiments the sequence \textit{Darkening} is used for training our models, which consists of 800 frames. The illumination of this scene is gradually changing from evening to night. For testing, the \textit{Light Switch} video is used, which comprises 600 frames. This sequence only has night scenes and features light-switch effects in the middle of the video, where a store light is suddenly switched off. Since this effect is not present in the training video, \textit{Light Switch} is an excellent candidate for measuring the generalisation abilities of our trained models.  The rest of the video sequence in the SABS dataset \cite{sabs} are not used because they are either day-time scenes and/or do not have significant illumination changes over time. Some example frames of the training and testing set are depicted in figure \ref{fig:SABS_overview}.

\subsection{Training parameters}
As mentioned above, we use the \textit{Darkening} video sequence for training the model and the \textit{Light Switch} video for testing.
All models are trained with the same parameters. The initial learning rate is $lr=0.001$ and is reduced by a factor of $0.1$ if the model does not improve for $2$ epochs. The training process ends after $5$ epochs of no improvements. For optimisation, the Adam optimiser \cite{adam14} is selected with betas $b_1=0.9$ and $b_2=0.999$.
Finally, the batch size is set to $1$.

To avoid overfitting, we freeze the encoder of our network. Specifically, the first 5 convolutional blocks of VGG16 are fixed and we only train the decoder. This training procedure yields better results according to our experiments.

Furthermore, for all augmenters, the probability of each training sample being augmented is set to 66.7\%.

Most frames contain many more pixels of the background than the foreground - some frames might not depict any moving objects at all. Given this observation, the loss function needs to balance the classes as to not allow the model to be biased towards the background class. Therefore, we use the weighted cross-entropy loss, which is formally defined as follows:
\begin{equation}
    G_s = wt [ - \log{\sigma{(x)}}] + (1 - t) [- \log{(1 - \sigma{(x)})}],
\end{equation}
where $w$ is the weight coefficient, $x$ is the predicted label, $t$ is the target label and $\sigma(x)=$ $1\over{1+e^{-x}}$ is a sigmoid function. The weight $w$ is calculated according to the ground truth frames with the following formula:
\begin{equation}
    w={N \over{2 \times [N_b,N_f]}},
\end{equation}

\noindent where N denotes the number of pixels of all input frames and $N_b$,$N_f$ are those pixels that belong to the background and foreground respectively.

\subsection{Implementation details}
We use the Keras library \cite{chollet2015keras} for training our models. In addition, for the quick deployment of the proposed model, the \textit{Segmentation models} \cite{segmentation_models_library} library is used. The full code is uploaded on GitHub \footnote{\url{https://github.com/dksakkos/illumination_augmentation}}. The Graphics Processing Unit (GPU) that was used in all our experiments is a GeForce GTX TITAN X.

\label{sec:ablation:local}
\begin{table*}
\centering
\begin{subtable}{\textwidth}
\centering
\begin{tabular}{ | c | c | c | c | c | c | c | c | c | c |}
\hline
Settings & Recall $\uparrow$ & Sp $\uparrow$ & FPR $\downarrow$ & FNR $\downarrow$ & PWC $\downarrow$ & FM $\uparrow$ & Precision $\uparrow$ & IoU $\uparrow$ & Matthews $\uparrow$ \\ \hline 
%vgg16_SABS_local_augment.h5 (t=0.7) \\ 
%z \in (80,120), d \in (1/2, 1/1.5) & 
$L_a$ &
\formatNumber{0.5467057122357221} &
\formatNumber{0.9962083107607284} &
\formatNumber{0.00379168923927167} &
\formatNumber{0.45329428776427794} &
\formatNumber{1.4289694444444445} &
\formatNumber{0.6411969077297167} &
\formatNumber{0.7751763801159087} &
\formatNumber{0.47188360946280095} &
\formatNumber{0.6442129151140162} \\ \hline

%vgg16_SABS_52_augment.h5 (t=0.6) \\(80,120), (1/5, 1/2) &
$L_b$ &
\formatNumber{0.59583420431099} &
\formatNumber{0.9951063775066578} &
\formatNumber{0.004893622493342271} &
\formatNumber{0.40416579568900995} &
\formatNumber{1.4218510416666668} &
\formatNumber{0.661863254865845} &
\formatNumber{0.7443505655424291} &
\formatNumber{0.4946155594878981} &
\formatNumber{0.658925410788799} \\ \hline

%vgg16_SABS_120_160_augment.h5 (t=0.6) \\(120,160), (1/5, 1/2)  &
$L_c$ &
\formatNumber{0.629440759494499} &
\formatNumber{0.9953573109013102} &
\formatNumber{0.00464268909868981} &
\formatNumber{0.37055924050550104} &
\formatNumber{1.3188565972222221} &
\formatNumber{0.6903320899516645} &
\formatNumber{0.7642662762643772} &
\formatNumber{0.5271046840616158} &
\formatNumber{0.6870217935725037} \\ \hline

\end{tabular}
\caption{Ablation studies for local changes}
\label{tbl:ablation_local}
\end{subtable}

\vspace{3mm}

\begin{subtable}{\textwidth}
\centering
\begin{tabular}{ | c | c | c | c | c | c | c | c | c | c |}
\hline
Settings & Recall $\uparrow$ & Sp $\uparrow$ & FPR $\downarrow$ & FNR $\downarrow$ & PWC $\downarrow$ & FM $\uparrow$ & Precision $\uparrow$ & IoU $\uparrow$ & Matthews $\uparrow$ \\ \hline 
%vgg16_SABS_local_augment.h5 (t=0.7) \\ 
%z \in (80,120), d \in (1/2, 1/1.5) & 
$G_{low}$ &
\formatNumber{0.7103310686427526} &
\formatNumber{0.9927183452771978} &
\formatNumber{0.00728165472280213} &
\formatNumber{0.2896689313572473} &
\formatNumber{1.3876729166666666} &
\formatNumber{0.7051011149861984} &
\formatNumber{0.6999476115956518} &
\formatNumber{0.5445221423437114} &
\formatNumber{0.6980158084633218} \\ \hline

%vgg16_SABS_glob_40_80_augment.h5 global changes (t=0.6) \\
%no circle used, all pixels changed evenly by z \in (40,80) &
$G_{med}$ &
\formatNumber{0.7082065274113394} &
\formatNumber{0.9942161276411903} &
\formatNumber{0.005783872358809675} &
\formatNumber{0.29179347258866056} &
\formatNumber{1.246354513888889} &
\formatNumber{0.7263373765554718} &
\formatNumber{0.7454209550915825} &
\formatNumber{0.5702745477378813} &
\formatNumber{0.7202118473013954} \\ \hline

$G_{high}$ &
\formatNumber{0.6678619503139831} &
\formatNumber{0.995240446141451} &
\formatNumber{0.004759553858548994} &
\formatNumber{0.3321380496860169} &
\formatNumber{1.2405385416666668} &
\formatNumber{0.7154782723095353} &
\formatNumber{0.7704056143812181} &
\formatNumber{0.5569997430841017} &
\formatNumber{0.711061065751801} \\ \hline

\end{tabular}

\caption{Ablation studies for global changes}
\label{tbl:ablation_global}
\end{subtable}

\caption{Ablation studies for local and global changes}
\label{tbl:ablation}
\end{table*}

\subsection{Evaluation Metric}
\label{sec:evaluation}
For evaluating our experiments, we use the following metrics: \textit{F-Measure (FM), Intersection over Union (IoU), Matthews correlation (MC)}. We provide the formal definitions below:
\begin{equation}
    Precision={TP\over TP+FP}
\end{equation}

\begin{equation}
    Recall={TP\over TP+FN}
\end{equation}

\begin{equation}
    FM={2 \times Precision \times Recall \over Precision+Recall}
\end{equation}

\begin{equation}
    IoU={TP \over TP + FP + FN}
\end{equation}

\begin{equation}
    %MC={TP \times TN - FP \times FN \over \sqrt{(TP+FP) \times (TP+FN) \times (TN+FP)
    %(TN+FP)\times (TN+FN)}}
    MC={TP \times TN - FP \times FN \over \sqrt{(TP+FP)(TP+FN)(TN+FP)(TN+FN)}} % from Wikipedia, is that right? I removed the \time between brackets
\end{equation}

where TP, TN, FP, FN denote the true positive, true negative, false positive and false negative pixels respectively.

\section{Results \label{sec:results}}
\begin{figure}
     \begin{subfigure}[c]{0.15\textwidth}
         \centering
         \includegraphics[width=\textwidth]{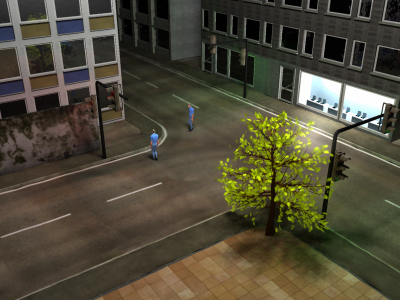}
         \label{fig:sabs_flip}
     \end{subfigure} %
     \hfill
     \begin{subfigure}[c]{0.15\textwidth}
         \centering
         \includegraphics[width=\textwidth]{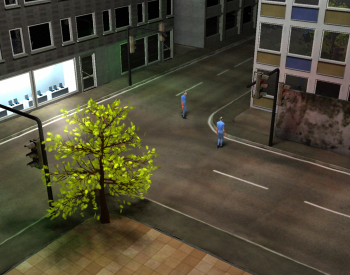}
         \label{fig:sabs_crop}
     \end{subfigure} %
    \hfill
     \begin{subfigure}[c]{0.15\textwidth}
         \centering
         \includegraphics[width=\textwidth]{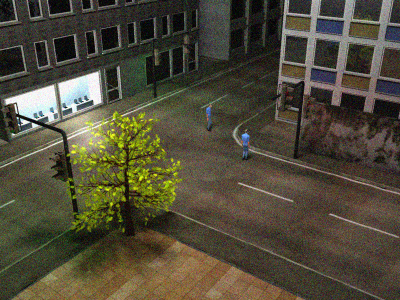}
         \label{fig:sabs_noise}
     \end{subfigure}
        \caption{Default augmentation techniques (from left to right): image mirroring, center cropping and adding noise.}
        \label{fig:regular_augmentation}
\end{figure}

We perform extensive evaluations on the proposed method. In particular, a wide range of different augmentation settings (Table \ref{tbl:settings}) were evaluated. 
We also compare against the regular augmentation techniques. We implement a "default" augmenter which performs the following image transformations: \textit{horizontal flipping, random cropping} and \textit{noise addition}, as depicted in Figure \ref{fig:regular_augmentation}. The \textit{cropping} operation performs center cropping with random image sizes, whereas the \textit{noise} option adds salt and pepper noise drawn from a Gaussian distribution. The amount of noise is fixed to $0.05$. All operations have a $50\%$ probability of taking place.

In the following section, we will evaluate the proposed method quantitatively to determine the optimal settings.

\begin{table*}
\begin{center}
\begin{tabular}{ |p{2cm}|p{10cm}|p{2cm}|}
\hline
\bf{Name} & \bf{Description} & \bf{Threshold} \\ \hline
baseline & No augmentation & 0.8 \\ \hline
default & Common augmentation: Mirror, crop and noise & 0.7 \\ \hline
$L_a$ & Local changes with $z \in (80, 120), k \in (1/2, 2/3) \times G$ & 0.7\\ \hline
$L_b$ & Local changes with $ z \in (80, 120), k \in (1/5, 1/2) \times G$ & 0.7 \\ \hline
$L_c$ & Local changes with $z \in (120, 160), k \in (1/5, 1/2) \times G$  & 0.6 \\ \hline
$G_{low}$ & Global, low intensity changes with $z \in (20, 60)$ & 0.9 \\ \hline
$G_{med}$ & Global, medium intensity changes with $z \in (40, 80)$ & 0.6 \\ \hline
$G_{high}$ & Global, high intensity changes with $z \in (60, 100)$ & 0.8 \\ \hline
GL & Global and local changes with $z_{global} \in (40, 80)$ and $z_{local} \in (120, 160)$ & 0.7 \\ \hline
%glob loc mild & local=(80,160), global=(40,80) 
%if local-global changes are both dark/bright
%then global=(0,200-local)
%(to avoid decreasing/increasing too much) 
%& 0.6 \\ \hline
%global \& local & combination of global and local changes
%local=(80,160), global=(40,200)
%if both are same, else global=(40+local, 200+local)  & 0.4 \\ \hline
%glob loc & with refine & 0.4\\ \hline
\end{tabular}
\end{center}
\caption{The different augmentation settings that were tested in our experiments. Parameters $k$, $z$ and $G$ denote the kernel size of the mask $M_1$, the illumination intensity in terms of pixel values and the resolution of the smallest dimension of the input image respectively. The last column shows the threshold that maximised the F-Measure of the segmentation mask.}
\label{tbl:settings}
\end{table*}

\subsection{Quantitative Evaluations}
In this experiment, we evaluate the performance of the proposed method using the commonly used metrics stated in section \ref{sec:evaluation} on the SABS dataset. The results of our experiments are presented in Table \ref{tabl:compareToNoAug}. Even though the default augmenter improved the F-Measure by more than 7\%, the proposed model, named $GL$, outperformed it by a very large margin of 16\%. As a matter of fact, the proposed method obtains better results in every single metric. This highlights the effectiveness of our proposed method. 

In addition, to determine the optimal settings, an ablation study is conducted and the details are explained in Section \ref{sec:ablation}.

\begin{table}
\begin{center}
\begin{tabular}{ | c | c | c | c | }
%$vgg16_SABS_no_augment.h5$ & (t=0.8)\\
\hline
Setting & No augm & Common augm & GL \\
\hline
Recall $\uparrow$ & \formatNumber{0.46061912313091663} &
\formatNumber{0.5439802335895172} &
\formatNumber{0.7687321647811425}\\ \hline
Sp $\uparrow$ & \formatNumber{0.9932679449090044} &
\formatNumber{0.9937371318019683} &
\formatNumber{0.9940734872885905}\\ \hline
FPR $\downarrow$ & \formatNumber{0.006732055090995628} &
\formatNumber{0.006262868198031724} &
\formatNumber{0.005926512711409545}\\ \hline
FNR $\downarrow$ & \formatNumber{0.5393808768690833} &
\formatNumber{0.45601976641048275} &
\formatNumber{0.2312678352188575}\\ \hline
PWC $\downarrow$ & \formatNumber{1.9171916666666666} &
\formatNumber{1.67668125} &
\formatNumber{1.1189295138888888}\\ \hline
FM $\uparrow$ & \formatNumber{0.5287970393345023} &
\formatNumber{0.6024543750220226} &
\formatNumber{0.762416869382313}\\ \hline
Precision $\uparrow$ & \formatNumber{0.620663743442177} &
\formatNumber{0.675013716502493} &
\formatNumber{0.7562044914382808}\\ \hline
IoU $\uparrow$ & \formatNumber{0.3594317395169605} &
\formatNumber{0.43108029122950176} &
\formatNumber{0.6160530557666739}\\ \hline
Matthews $\uparrow$ & \formatNumber{0.5252549314185203} &
\formatNumber{0.5976034982281958} &
\formatNumber{0.7567144790377998}\\ \hline
\end{tabular}
\end{center}
\caption{Comparison between no augmentation, common augmentation and the proposed method which covers global and local illumination changes.}
\label{tabl:compareToNoAug}
\end{table}

\subsection{Qualitative Evaluations}

\begin{figure*}
\begin{center}
\begin{tabular}{ c c c c }
         \rotatebox{90}{input frames} & 
         \includegraphics[width=0.25\textwidth]{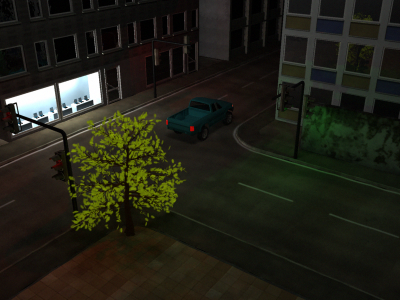} &
         \includegraphics[width=0.25\textwidth]{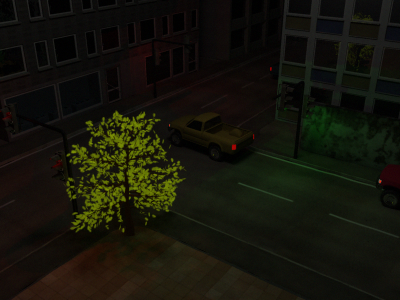} &
         \includegraphics[width=0.25\textwidth]{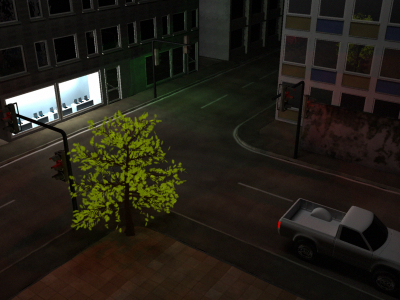} \\
         \rotatebox{90}{no augmentation} & 
         \includegraphics[width=0.25\textwidth]{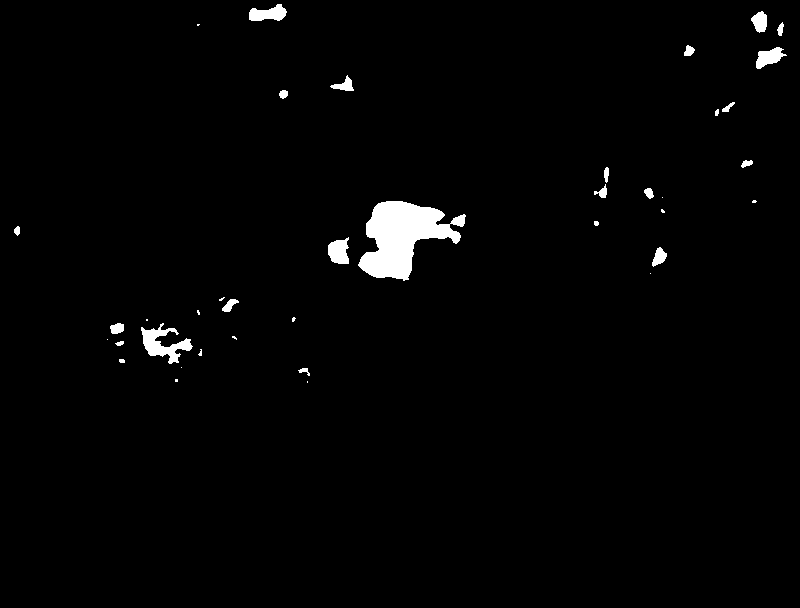} &
         \includegraphics[width=0.25\textwidth]{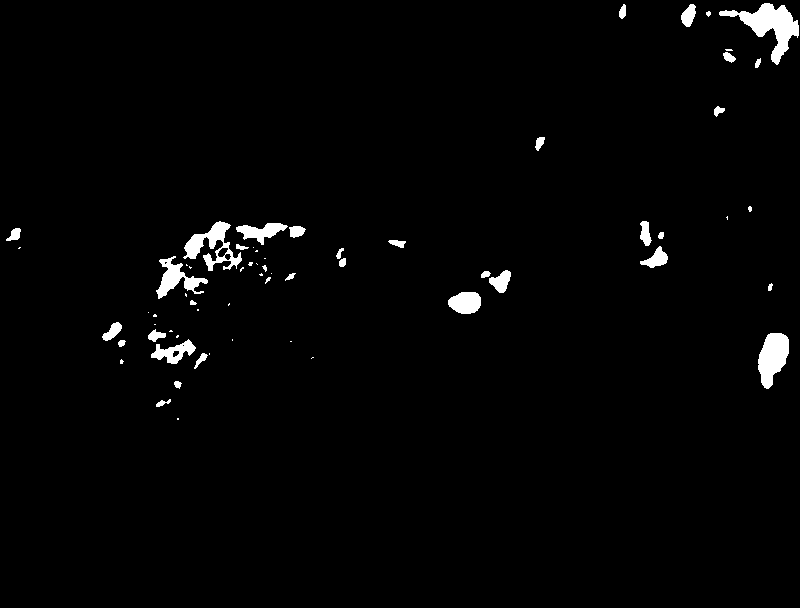} &
         \includegraphics[width=0.25\textwidth]{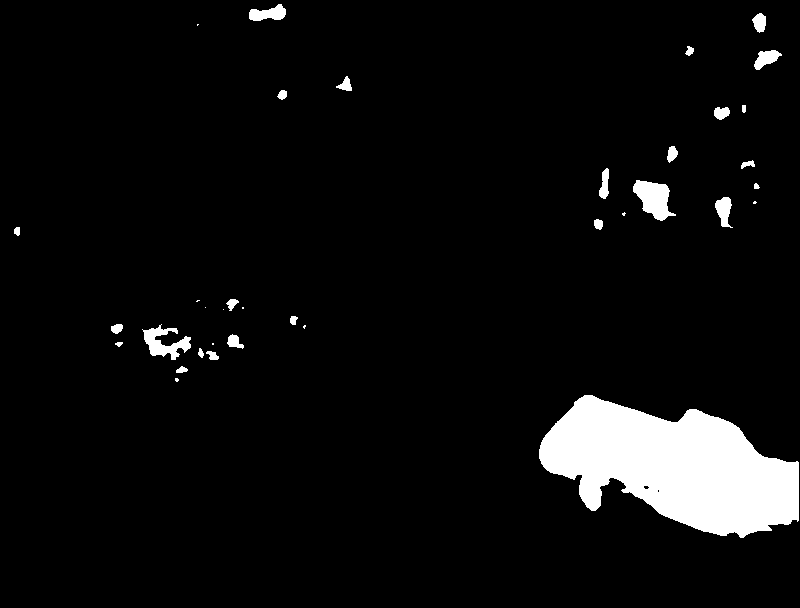} \\
         \rotatebox{90}{default augmentation} & 
         \includegraphics[width=0.25\textwidth]{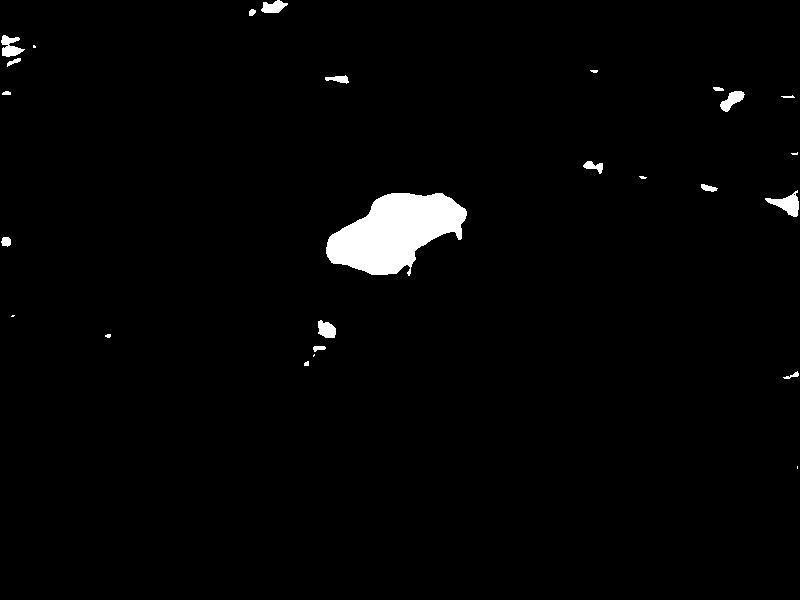} &
         \includegraphics[width=0.25\textwidth]{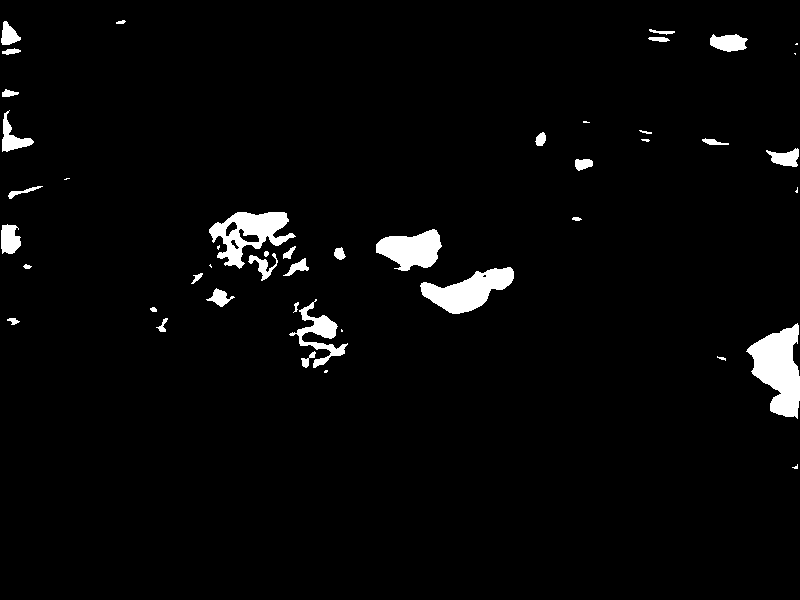} &
         \includegraphics[width=0.25\textwidth]{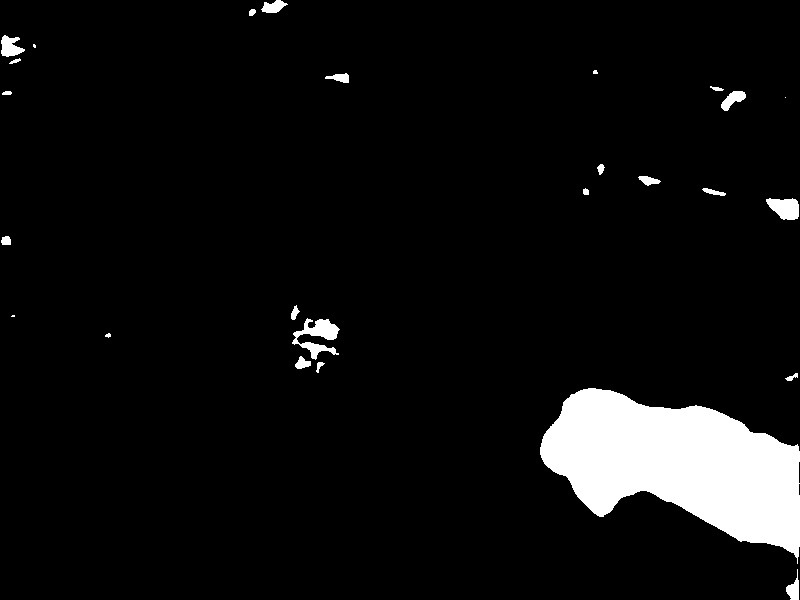} \\
         \rotatebox{90}{local changes} & 
         \includegraphics[width=0.25\textwidth]{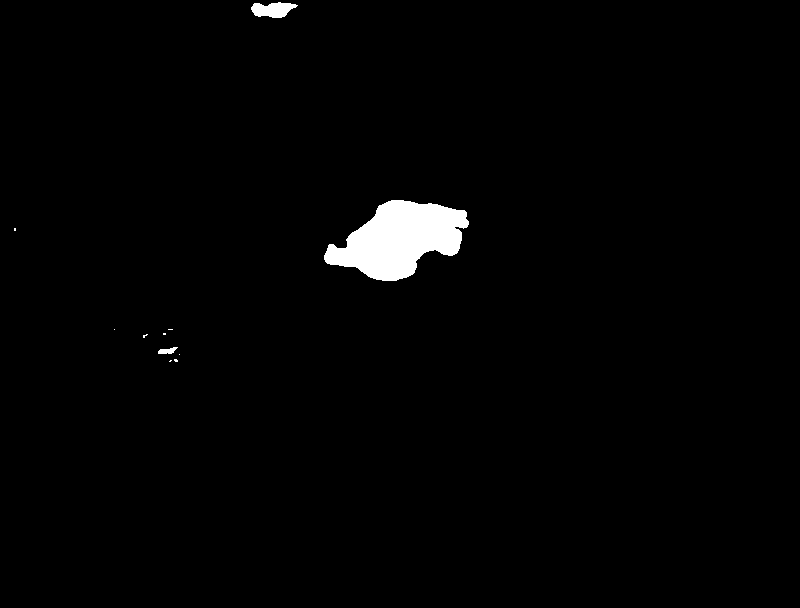} &
         \includegraphics[width=0.25\textwidth]{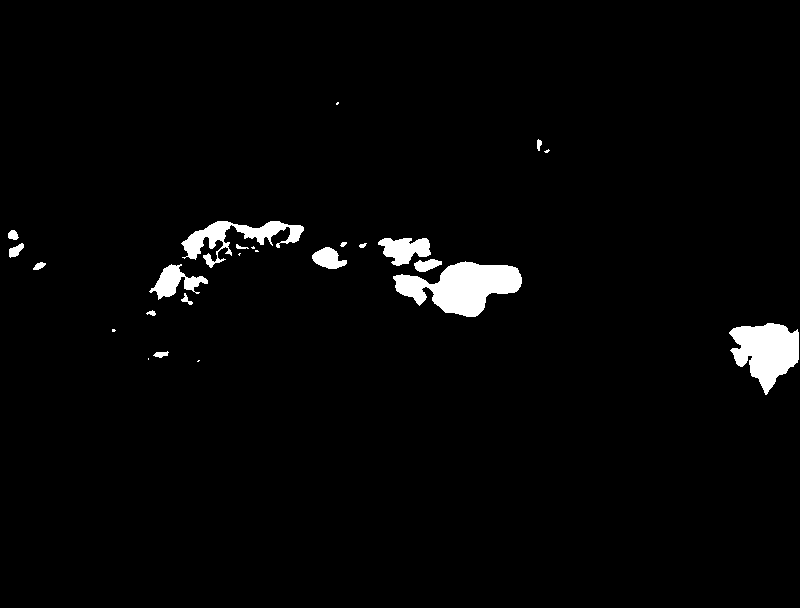} &
         \includegraphics[width=0.25\textwidth]{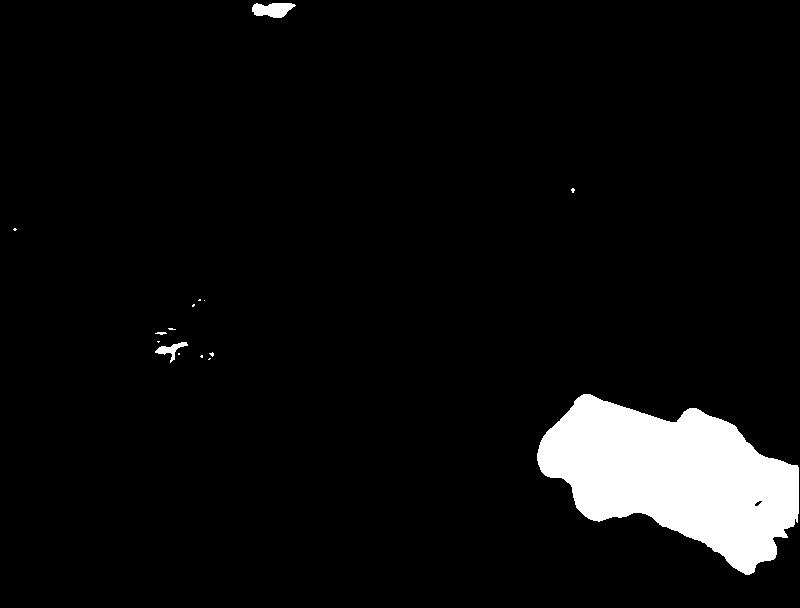} \\
         \rotatebox{90}{global changes} & 
         \includegraphics[width=0.25\textwidth]{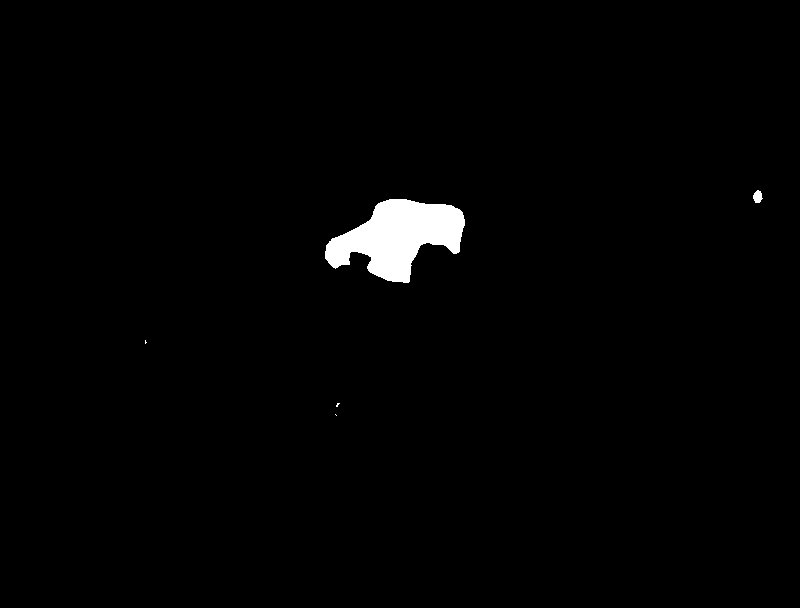} &
         \includegraphics[width=0.25\textwidth]{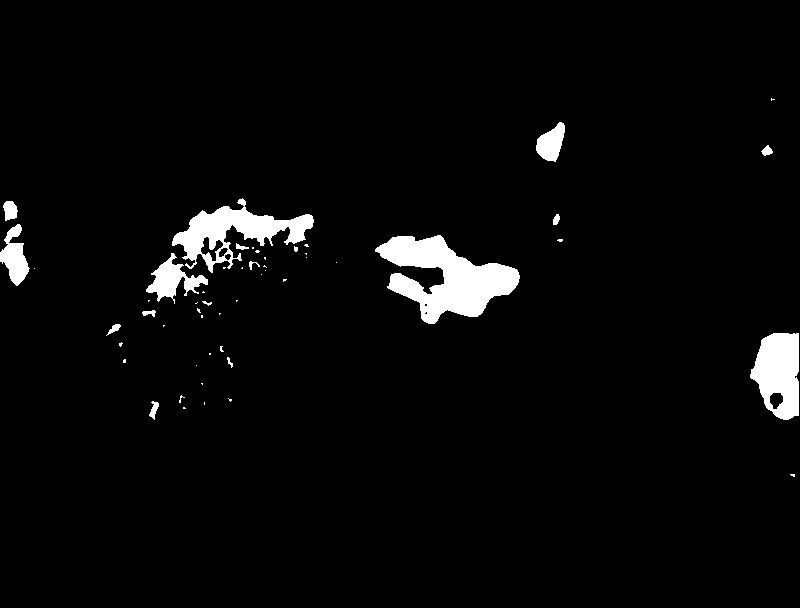} &
         \includegraphics[width=0.25\textwidth]{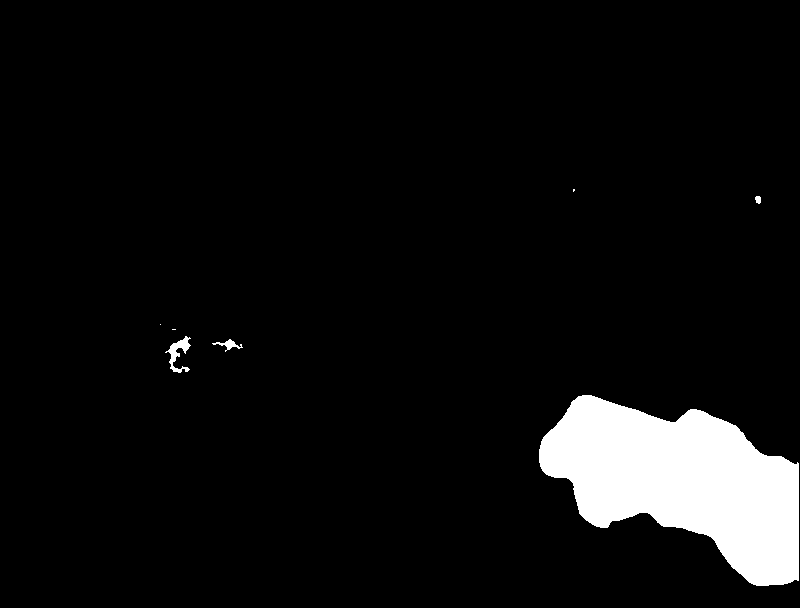} \\
         \rotatebox{90}{global and local} & 
         \includegraphics[width=0.25\textwidth]{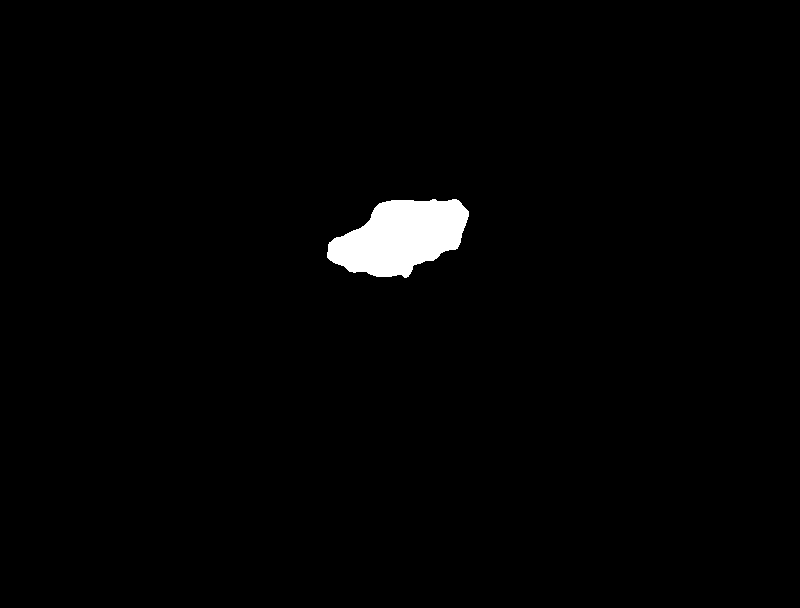} &
         \includegraphics[width=0.25\textwidth]{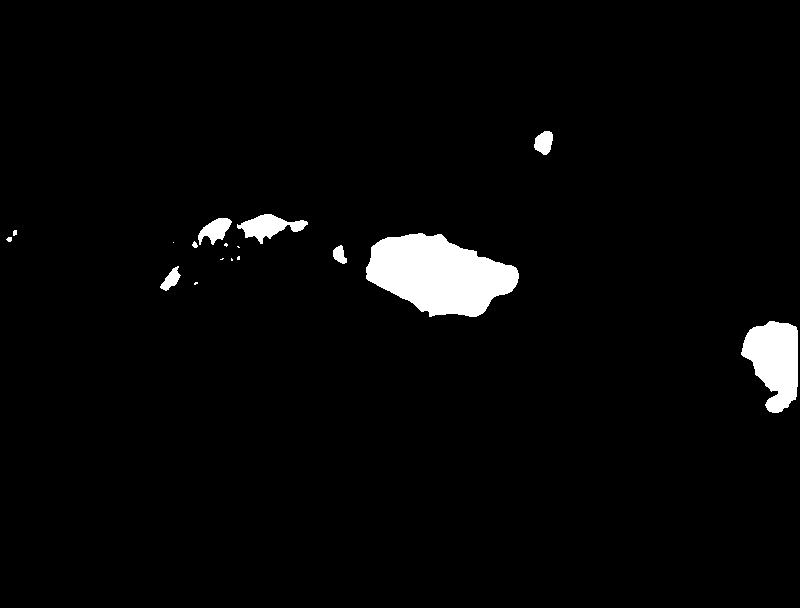} &
         \includegraphics[width=0.25\textwidth]{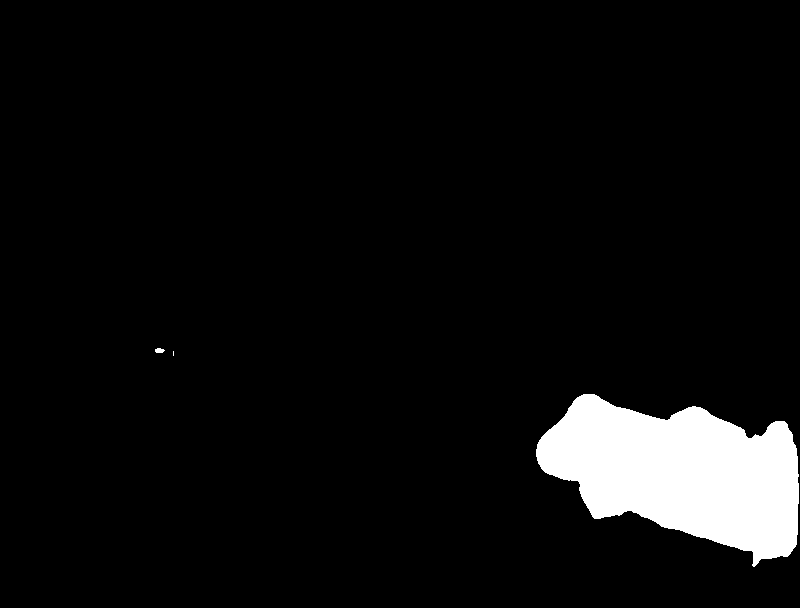} \\
         \rotatebox{90}{ground truths} & 
         \includegraphics[width=0.25\textwidth]{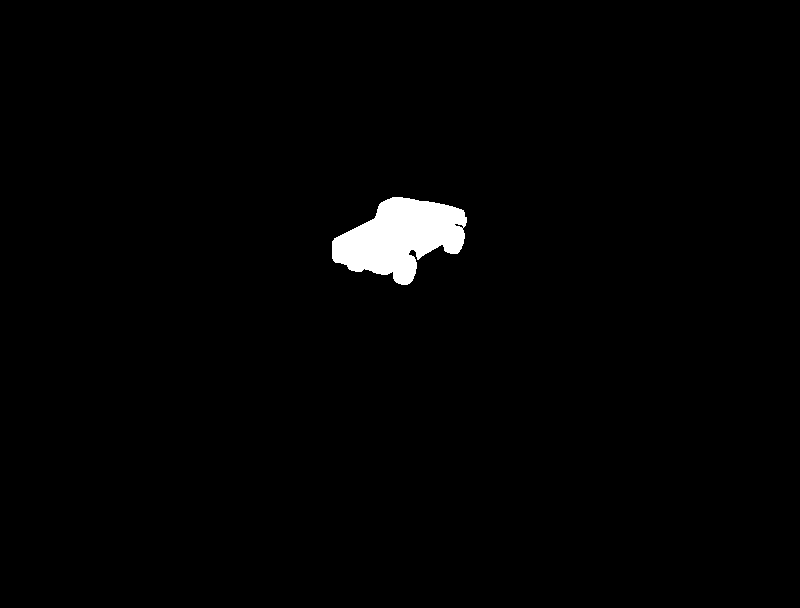} &
         \includegraphics[width=0.25\textwidth]{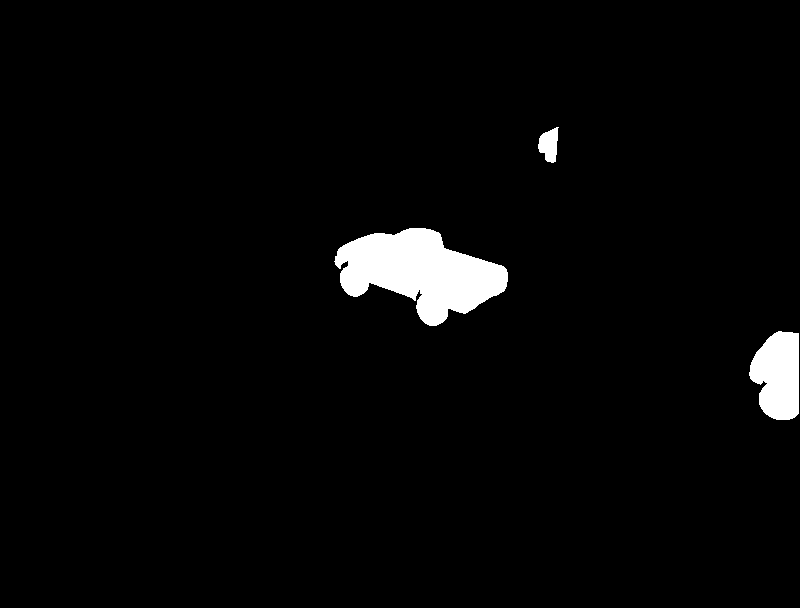} &
         \includegraphics[width=0.25\textwidth]{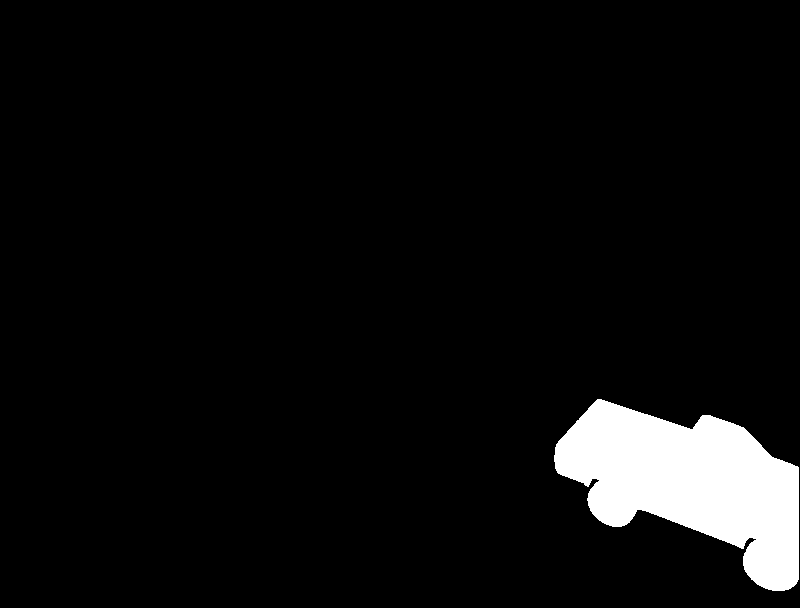} \\
\end{tabular}
\end{center}
\caption{Comparison between different augmentation techniques.}
\label{fig:different_aug}
\end{figure*}

To provide visual comparison between the various models tested in this paper, we report the segmentation results of three frames taken from the start, middle and end of the video.
The results are illustrated in Figure \ref{fig:different_aug}. The comparison indicates that the proposed data augmentation approach improves the quality of the segmentation masks, as it obtained solutions which are clearly closer to the ground truths compared to other models.

\begin{figure*}
     \begin{subfigure}[b]{0.32\textwidth}
         \centering
         \includegraphics[width=\textwidth]{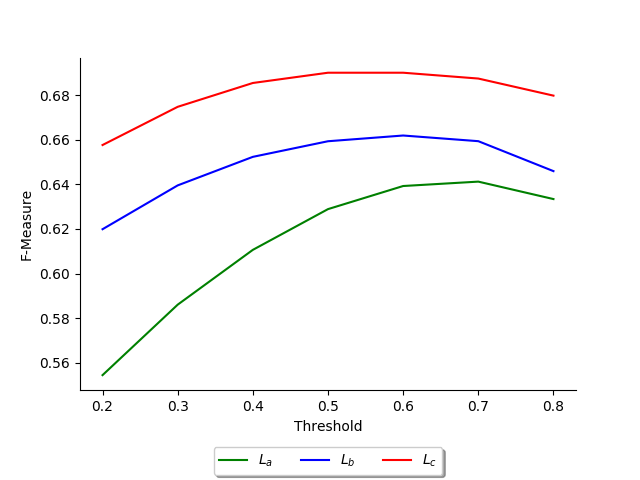}
     \end{subfigure}
     \hfill
        \begin{subfigure}[b]{0.32\textwidth}
         \centering
         \includegraphics[width=\textwidth]{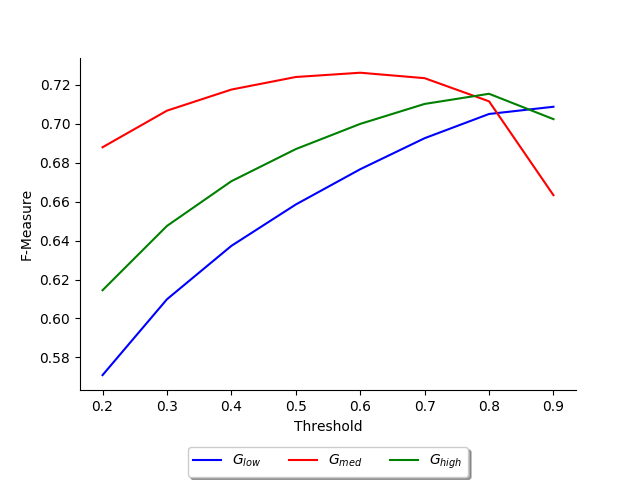}
     \end{subfigure}
     \hfill
     \begin{subfigure}[b]{0.32\textwidth}
         \centering
         \includegraphics[width=\textwidth]{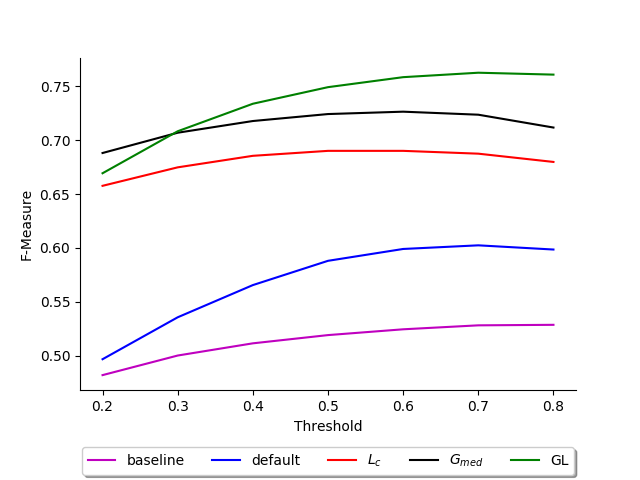}
     \end{subfigure}
        \caption{F-Measure values on different thresholds for each model.}
        \label{fig:fm-thresholds}
\end{figure*}

\subsection{Ablation studies}
\label{sec:ablation}
Here we discuss the ablation studies that verify the optimal hyper-parameters of our method. The full list of our experiments can be found in Table \ref{tbl:settings}, whereas the result of each method is shown in Table \ref{tbl:ablation}. By cross-checking these tables, it can be seen that using a smaller kernel is better for creating local effects. Also, greater changes in illumination yield better results. This is because of the fading effect caused by the distance transform, which is only strong in the centre of the circle.
For global changes a much smaller noise value is needed, firstly because there is no fading and secondly due to the effect being applied to the whole image.

It is evident from Table \ref{tbl:ablation} that both local and global change augmentations yield significant results. However, the best performing model according to our experiments encompasses both.

It is noteworthy that sub-optimal settings need a very high threshold to produce a good segmentation result. This is because in those frames of the \textit{Light Switch} sequence where the light suddenly switches off, the model fails to identify the moving object due to low lighting. However, with the optimal settings of the proposed method, the model can generalise in all illumination conditions. The performance of each model under different threshold is depicted in Figure \ref{fig:fm-thresholds}.

\section{Conclusion and Future Work}
\label{sec:conclusion}
In this paper, we have presented a fast and easy method to synthesise training samples for the implementation of illumination-invariant models. The synthetic images are generated by artificially altering the pixel intensity values of the input image not only globally but also in small regions. A typical "lamp-post" light source effect can be approximated by applying the distance transform on a binary mask. 
We have tested the proposed method in the task of background subtraction. The experimental results indicate that the models trained using the dataset augmented with the new synthetics are more robust to illumination changes and are able to handle even intense lighting variations. 
As future work, more shapes can be explored, not only geometrical but also of arbitrary shapes, for representing shadows more realistically. 

\bibliographystyle{IEEEtran}
\bibliography{SKIMA2019}

\end{document}